\documentclass[conference]{IEEEtran}
\IEEEoverridecommandlockouts
\usepackage{cite}
\usepackage{amsmath,amssymb,amsfonts}
\usepackage{multirow}
\usepackage{algorithm}
\usepackage{subcaption}

\usepackage{algpseudocode}
\usepackage{amsmath,amssymb,amsfonts}
\usepackage{graphicx}
\usepackage{textcomp}
\usepackage{xcolor}
\def\BibTeX{{\rm B\kern-.05em{\sc i\kern-.025em b}\kern-.08em
    T\kern-.1667em\lower.7ex\hbox{E}\kern-.125emX}}
\begin{document}

\title{Visual Bias and Interpretability in Deep Learning for Dermatological Image Analysis}


\author{\IEEEauthorblockN{Enam Ahmed Taufik$^{1}$, Abdullah Khondoker$^{2}$, Antara Firoz Parsa$^{2}$, Seraj Al Mahmud Mostafa$^{3}$}
\IEEEauthorblockA{\textit{$^1$Department of Computer Science and Engineering, European University of Bangladesh,} Dhaka, Bangladesh\\
\textit{$^2$Department of Computer Science and Engineering, BRAC University,} Dhaka, Bangladesh\\
\textit{$^3$Department of Information Systems, University of Maryland, Baltimore County,} Baltimore, MD, USA\\
taufik@eub.edu.bd$^{1}$, ext.abdullah@bracu.ac.bd$^{2}$, antara.firuz.parsa@g.bracu.ac.bd$^{2}$,  serajmostafa@umbc.edu$^{3}$}
}







\maketitle

\begin{abstract}
Accurate skin disease classification is a critical yet challenging task due to high inter-class similarity, intra-class variability, and complex lesion textures. While deep learning-based computer-aided diagnosis (CAD) systems have shown promise in automating dermatological assessments, their performance is highly dependent on image pre-processing and model architecture. This study proposes a deep learning framework for multi-class skin disease classification, systematically evaluating three image pre-processing techniques: standard RGB, CMY color space transformation, and Contrast Limited Adaptive Histogram Equalization (CLAHE). We benchmark the performance of pre-trained convolutional neural networks (DenseNet201, EfficientNetB5) and transformer-based models (ViT, Swin Transformer, DinoV2 Large) using accuracy and F1-score as evaluation metrics. Results show that DinoV2 with RGB pre-processing achieves the highest accuracy (up to 93\%) and F1-scores across all variants. Grad-CAM visualizations applied to RGB inputs further reveal precise lesion localization, enhancing interpretability. These findings underscore the importance of effective pre-processing and model choice in building robust and explainable CAD systems for dermatology.
\end{abstract}


\section{INTRODUCTION}

Skin diseases encompass a wide range of conditions, including inflammatory, infectious, and neoplastic disorders, that pose diagnostic challenges due to visual similarities, intra-class variability, and complex morphology. Early and accurate detection, particularly of malignant cancers and viral infections, is vital for effective treatment and containment~\cite{ref1,ref2}. Traditional diagnostic methods such as visual inspection and dermoscopy are subjective and suffer from inter-observer variability, with dermatologists achieving only 60-80\% accuracy~\cite{ref3,ref4}. Viral infections like chickenpox, measles, and monkeypox further complicate diagnosis due to overlapping features~\cite{ref5,ref6}.

Computer-aided diagnosis (CAD) systems aim to address these limitations using machine learning and deep learning techniques~\cite{ref7}. However, traditional models with handcrafted features (e.g., SVMs, Random Forests) struggle with high-dimensional image data~\cite{ref8}.
Deep learning, particularly CNNs, has shown superior performance by learning hierarchical features from images~\cite{ref9}. More recently, vision transformers (ViTs) have demonstrated strong classification capabilities by capturing long-range dependencies~\cite{ref10}. Hybrid models combining CNNs, transformers, and attention mechanisms further enhance feature representation in complex multi-class settings~\cite{ref11, taufik2025efficient}. Still, their success relies on effective pre-processing, augmentation, and interpretability strategies.

In this study, we present a deep learning-based framework for multi-class skin disease classification and conduct a comprehensive analysis of how different image pre-processing techniques and model architectures influence performance and interpretability. We investigate three pre-processing strategies: (i) standard RGB images, (ii) CMY color space transformation, and (iii) Contrast Limited Adaptive Histogram Equalization (CLAHE). These are evaluated using two families of model architectures: pre-trained CNN-based transfer learning models and transformer-based models. Our experiments reveal that pre-processing strategies have a significant impact on classification accuracy, with CLAHE-enhanced images consistently improving performance across all architectures. Furthermore, transformer-based models outperformed CNNs in scenarios involving high inter-class visual similarity. Gradient-weighted Class Activation Mapping (Grad-CAM) visualizations indicate that models trained on CLAHE images focus more precisely on lesion regions, enhancing explainability. These findings emphasize the importance of thoughtful pre-processing and model selection in building fair, accurate, and explainable CAD systems for dermatological diagnosis.

The remainder of this paper is organized as follows. Related works in Section~\ref{sec:rw} reviews recent advancements in image classification, including optimization strategies, hybrid models, and dataset integration. Section~\ref{sec:meth} details the methodology, covering pre-processing, model design, and unified training across datasets. Experiments details are in Section~\ref{sec:exp}. Section~\ref{sec:res} represents results and analysis of classification performance, comparative analysis, and key insights. Finally, Section~\ref{sec:conc} concludes the work by outlining future research directions.

\section{RELATED WORK}
\label{sec:rw}
Recent advancements in deep learning have significantly enhanced skin disease classification systems. Jeyageetha et al.~\cite{ref13} proposed a model integrating Region Growing segmentation and MobileSkinNetV2, optimized using the Modified Honey Badger Optimizer (MHBO), achieving 99.01\% accuracy and 98.6\% precision on the ISIC dataset. Qasim et al.~\cite{ref14} introduced a hybrid framework combining Deep Spiking Neural Networks (DSNN) and VGG-13, reaching 89.5\% accuracy and a 90\% F-measure on ISIC 2019. Abbas et al.~\cite{ref15} evaluated CNN variants including DenseNet-121 and ResNet-50 on HAM10000, with their sequential CNN achieving AUC scores of up to 99\%. Helal et al. used geometric mean to address data imbalanceness \cite{al2016algorithms}. To address class imbalance, they used image augmentation techniques. Similarly, Mostafa et al. used more augmentation techniques to tackle challenges in complex image processing as well as custom model building tasks \cite{mostafa2025gwavenet, mostafa2025enhancing, mostafa2025multiscale}. Sasithradevi et al.~\cite{ref16} proposed EffiCAT, a hybrid model of EfficientNet and attention mechanisms (DCAL, CBAM), attaining 94.48\% accuracy on HAM10000 and 92.08\% on ISIC 2018. Aditya et al.~\cite{ref17} and Gonzalez et al.~\cite{gonzalez2022atmospheric} demonstrated accuracy gains using transfer learning techniques over training. CAD-Skin by Khan et al.~\cite{ref18} integrated CNNs, autoencoders, and Quantum SVMs, achieving up to 99\% accuracy across PAD-UFES-20, ISIC-2018, and ISIC-2019 datasets. Xiao et al.~\cite{ref19} analyzed ColorJitter augmentation on SID data, finding brightness and hue effective, though limitations persisted due to data quality. Lastly, Chu et al.~\cite{ref20} used UAV-based multispectral imaging with YOLOv5 and SSD, achieving 0.93 and 0.92 accuracy, highlighting deep learning's utility in forestry.

\section{METHODOLOGY}
\label{sec:meth}
\begin{figure}[ht]
    \centering
    \includegraphics[width=1\linewidth]{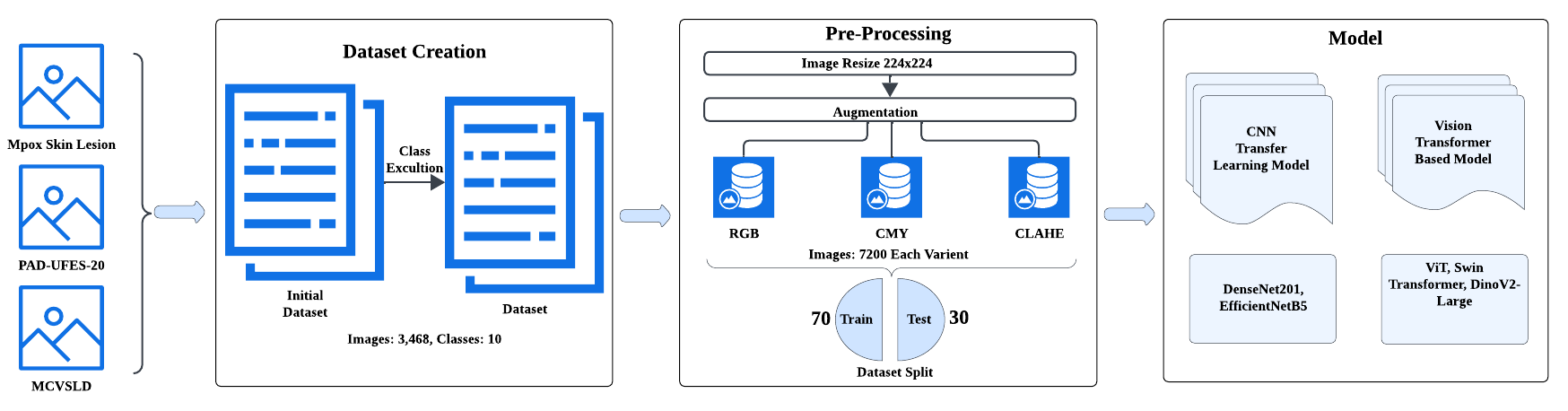}
    \caption{Methodology Pipeline}
    \label{fig:method_diagram}
\end{figure}

The overall methodology, illustrated in (Figure~\ref{fig:method_diagram}), outlines the complete pipeline from dataset preparation and pre-processing to model training.

\subsection{DATASET}
\subsubsection{DATASET OVERVIEW}
To ensure robustness and generalizability in multi-class skin disease classification, we curated a comprehensive dataset by integrating three publicly available sources: the \textbf{Mpox Skin Lesion~\cite{Ali2024}, PAD-UFES-20~\cite{pacheco2020pad}, and the Multi-Class Viral Skin Lesion Dataset (MCVSLD)~\cite{ahmed2025multi}}. This multi-source combination captures a wide range of dermatological conditions and visual variability across diverse patient populations and imaging environments.

The initial aggregated dataset consisted of 4,312 clinical skin images spanning 12 categories: \textit{Actinic Keratosis}, \textit{Basal Cell Carcinoma}, \textit{Chickenpox}, \textit{Measles}, \textit{Hand, Foot, and Mouth Disease (HFMD)}, \textit{Monkeypox}, \textit{Nevus}, \textit{Normal} (healthy skin), \textit{Seborrheic Keratosis}, \textit{Squamous Cell Carcinoma}, \textit{Melanoma}, and \textit{Cowpox}. Due to severe class imbalance and insufficient representation, the \textit{Melanoma} and \textit{Cowpox} classes were excluded. After this refinement, the final dataset comprised 3,468 images across 10 well-represented classes: nine skin diseases and one healthy skin category. Among these, \textit{Measles} is the most underrepresented with 147 samples, while \textit{Basal Cell Carcinoma} has the highest number of images, totaling 690.

\subsubsection{DATASET PRE-PROCESSING}
To enhance model robustness and mitigate class imbalance, an augmentation pipeline was applied using the Albumentations library. The transformations included horizontal and vertical flips, 90-degree rotations, brightness/contrast adjustments, shift-scale-rotate operations, and elastic distortions. For underrepresented classes (less than 500 samples), data was augmented to 600 images; for others, to 1,000 images.

\begin{figure}[ht]
\centering
\subfloat[Original]{\label{subfig:orig} \includegraphics[width=0.115\textwidth]{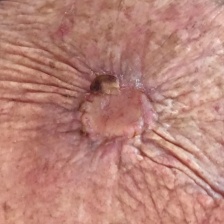}}
\hfill
\subfloat[RGB]{\label{subfig:RGB} \includegraphics[width=0.115\textwidth]{Images/Basal_Cell_Carcinoma_rgb.jpg}}%
\hfill
\subfloat[CMY]{\label{subfig:CMY} \includegraphics[width=0.115\textwidth]{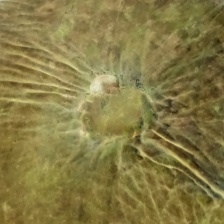}}%
\hfill
\subfloat[CLAHE]{\label{subfig:CLAHE} \includegraphics[width=0.115\textwidth]{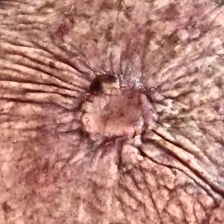}}%
\caption{Original and preprocessed images using various approaches: (a) Original, (b) Resized - RGB, (c) Resized - CMY, and (d) Resized - CLAHE.}
\label{fig:prep}
\end{figure}

Following augmentation, three pre-processing strategies were applied: (i) RGB-based augmentation, (ii) CMY color space transformation to enhance channel separation, and (iii) CLAHE (Contrast Limited Adaptive Histogram Equalization) to improve local contrast. These resulted in four dataset variants used in experimentation: original RGB, augmented RGB, CMY, and CLAHE-enhanced images (Figure~\ref{fig:prep}). After augmentation, the total number of images increased to 7,200. The dataset was then split into training and testing subsets using a 70:30 ratio.

\subsection{MODELS ARCHITECTURE}

\subsubsection{TRANSFER LEARNING MODELS}

We employed transfer learning with two high-capacity CNN backbones: \textbf{DenseNet201~\cite{huang2017densenet}}, and \textbf{EfficientNetB5~\cite{tan2019efficientnet}}, all pretrained on ImageNet. Each model was adapted by removing the top classification layer and unfreezing the last 30 layers for fine-tuning. A unified architecture was built around each backbone. An initial Conv2D layer (3×3, ReLU) normalized input channels, followed by the frozen base model. Post-feature extraction, we added a Conv2D layer (256 filters), Batch Normalization, Dropout (0.5), and Global Average Pooling. Two fully connected layers (1024 and 512 units) with ReLU activation, $L_2$ regularization (with penalty $\lambda = 10^{-4}$), Batch Normalization, and Dropout (0.6) were stacked before the final Dense softmax layer for 10-class classification, outputting $\hat{y} \in \mathbb{R}^{10}$. Models were trained using the Adam optimizer and an exponential learning rate decay:
\[
\eta_t = \eta_0 \cdot \gamma^{t}
\]
where $\eta_0 = 10^{-4}$ and $\gamma = 0.9$. We used categorical cross-entropy loss and monitored accuracy. Each model was trained for up to 100 epochs with early stopping and learning rate reduction on the plateau. The training pipeline utilized the \texttt{tf.data} API for efficient batching, shuffling, and prefetching.

\subsubsection{TRANSFORMER MODELS}
We evaluated three state-of-the-art vision transformer backbones: \textbf{Vision Transformer (ViT)}~\cite{dosovitskiy2020image}, \textbf{Swin Transformer}~\cite{liu2021swin}, and \textbf{DINOv2-L}~\cite{oquab2023dinov2}, all pretrained on ImageNet. For each model, the final classification head was replaced with a task-specific linear layer $f: \mathbb{R}^d \rightarrow \mathbb{R}^{10}$, producing logits $\hat{y} \in \mathbb{R}^{10}$ over the 10 skin disease classes. The models were fine-tuned end-to-end using PyTorch.

Images were loaded via a custom dataset class and augmented using standard transforms. Training was conducted using the Adam optimizer with a learning rate $\eta = 10^{-4}$ and CrossEntropyLoss:
\[
\mathcal{L}_{\text{CE}} = -\sum_{i=1}^{10} y_i \log(\hat{y}_i)
\]
A StepLR scheduler (decay factor 0.1 every 10 epochs) managed $\eta$, and gradient clipping with max-norm $||\nabla||_2 \leq 1.0$ was applied. Each model was trained for up to 50 epochs with early stopping (patience = 5) based on validation accuracy. The best-performing checkpoint was saved for evaluation. Training and validation were executed on a CUDA-enabled GPU, and performance was monitored via accuracy and loss metrics.

\begin{algorithm}[ht]
\caption{Pipeline for Skin Disease Classification}
\label{alg:method}
\begin{algorithmic}[1]
\Require Datasets: MSLD v2.0, PAD-UFES-20, MCVSLD
\Ensure Trained model with optimal test accuracy

\State \textbf{Data Curation:} Merge datasets; exclude \textit{Melanoma} and \textit{Cowpox}

\For{each class}
    \If{samples $<$ 500}
        \State Augment to 600
    \Else
        \State Augment to 1,000
    \EndIf
\EndFor

\State Generate 4 variants: RGB, Augmented RGB, Augmented CMY, Augmented CLAHE

\State \textbf{Split:} 70:30  train-test split for each image set.

\For{CNN model in \{DenseNet201, EfficientNetB5\}}
    \State Load ImageNet weights; unfreeze last 30 layers
    \State Append custom Conv, FC, Dropout layers
\EndFor

\For{Transformer in \{ViT, Swin, DINOv2-L\}}
    \State Replace head; fine-tune with Adam optimizer
\EndFor

\State Train with early stopping, save the best model

\end{algorithmic}
\end{algorithm}

\section{EXPERIMENTS}
\label{sec:exp}
\begin{figure}[ht]
\subfloat[RGB]{\label{subfig:RGB} \includegraphics[width=0.15\textwidth]{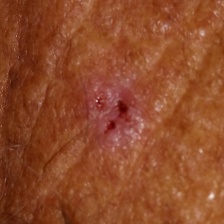}}%
\hfill
\subfloat[CMY]{\label{subfig:CMY} \includegraphics[width=0.15\textwidth]{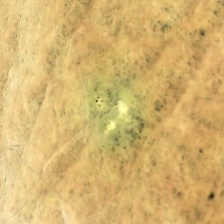}}%
\hfill
\subfloat[CLAHE]{\label{subfig:CLAHE} \includegraphics[width=0.15\textwidth]{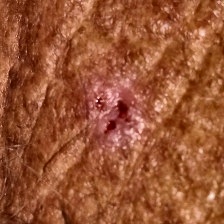}}%
\hfill
\caption{Comparison of lesion boundaries using (a) RGB, (b) CMY, and (c) CLAHE pre-processing.}
\label{fig:img}
\end{figure}

To assess the impact of different pre-processing techniques on visual quality and model performance, we compared lesion representations in RGB, CMY, and CLAHE-enhanced images (Figure~\ref{fig:img}). Among these, CMY-transformed images consistently yielded sharper boundary contrasts and improved lesion delineation. This enhancement is attributed to the subtractive nature of the CMY color model, which promotes chromatic separation and suppresses background interference~\cite{cheng2001color}. Despite perceptual advantages, transfer learning models trained on CMY inputs slightly underperformed relative to RGB-based models, primarily due to the domain-specific bias of pretrained feature extractors optimized on large-scale RGB datasets such as ImageNet~\cite{deng2009imagenet}. This domain shift~\cite{tzeng2017adversarial} limits transfer learning effectiveness for CMY images. In contrast, transformer-based models exhibited greater robustness across pre-processing strategies. Results from RGB, CMY, and CLAHE inputs were nearly equivalent in classification performance, suggesting that transformers are less susceptible to input distribution shifts and training bias, likely due to their patch-wise tokenization and global self-attention mechanisms that enable better generalization across color domains. 
Moreover, CMY pre-processing offers contrast enhancement comparable to CLAHE~\cite{pisano1998contrast}, a widely used medical imaging technique, but with substantially lower computational cost. While CLAHE performs localized histogram equalization and interpolation with complexity $\mathcal{O}(N \log L)$, CMY transformation is a simple linear inversion, $\text{CMY}(x,y) = 255 - \text{RGB}(x,y)$, with linear complexity $\mathcal{O}(N)$, making it more suitable for resource-constrained scenarios.

\begin{figure} [ht]
\hfill
\subfloat[CMY]{\label{subfig:org} \includegraphics[width=0.25\textwidth]{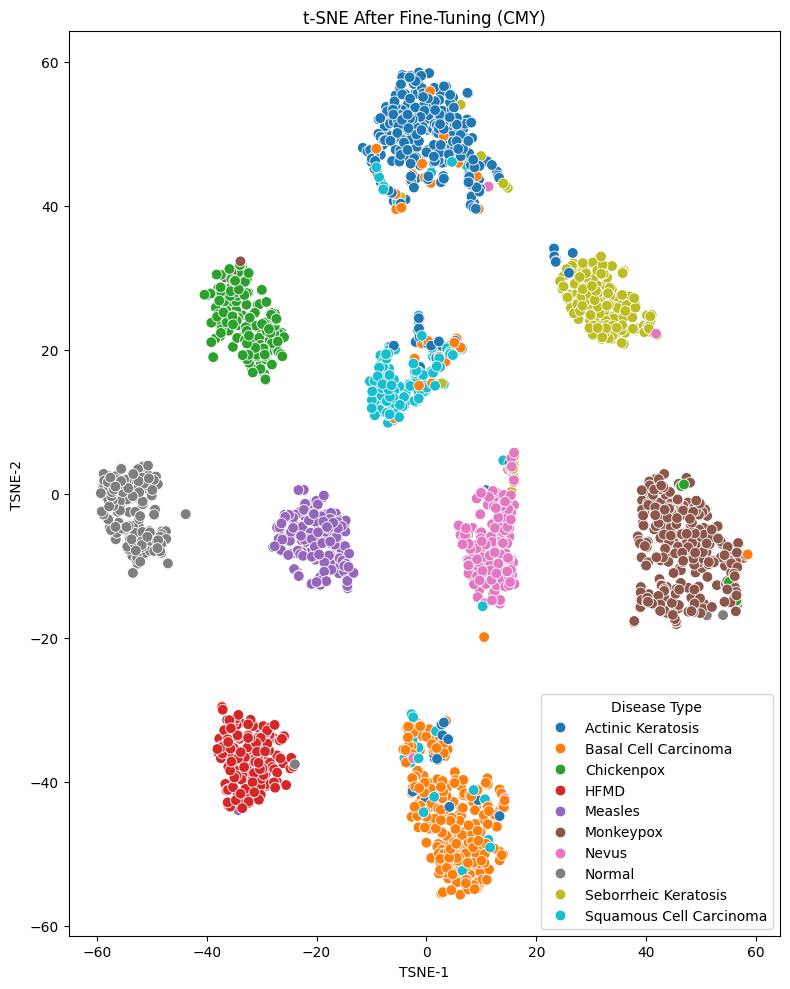}}%
\hfill
\subfloat[CLAHE]{\label{subfig:heatmap} \includegraphics[width=0.21\textwidth]{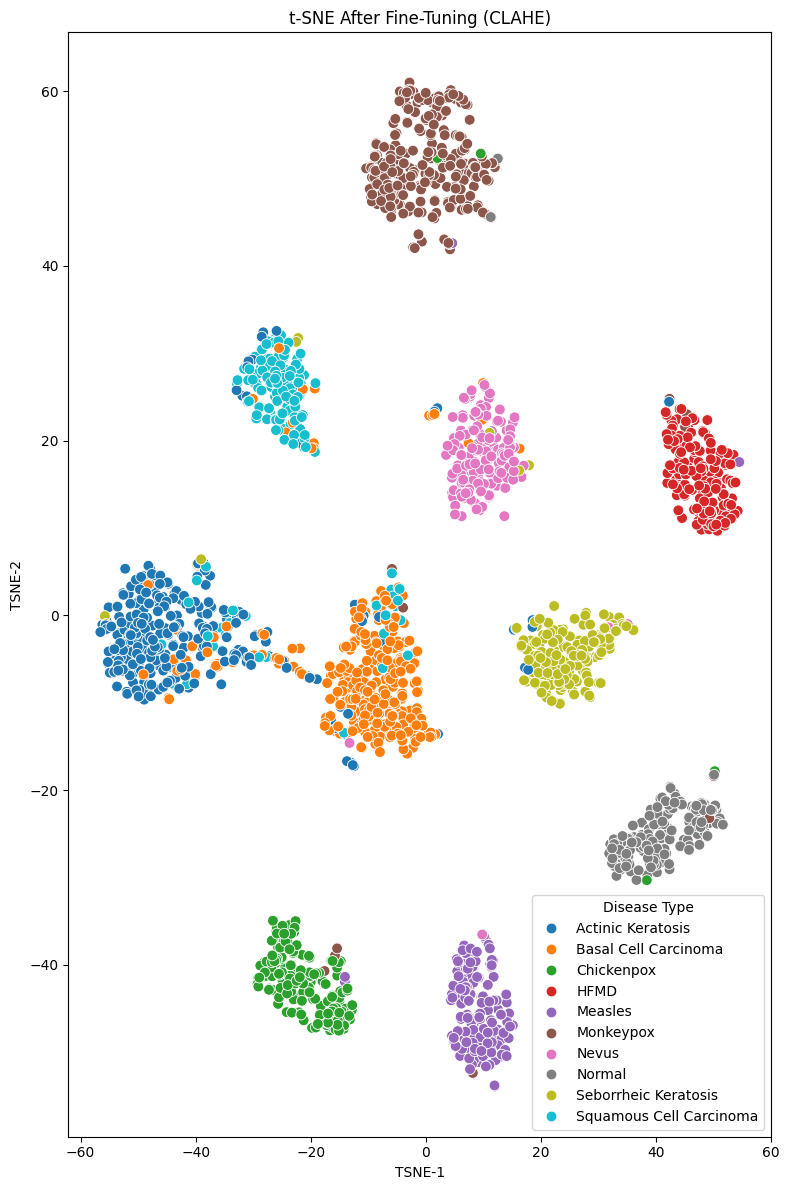}}%
\hfill
\caption{t-SNE of Fine-tuned DinoV2 Large Model for (a) CMY and (b) CLAHE}
\label{fig:tsne}
\end{figure}

We further fine-tuned the proposed DinoV2 Large model on RGB, CLAHE, and CMY-preprocessed images, visualizing feature embeddings via t-SNE (Figure~\ref{fig:tsne}). DinoV2 Large demonstrated comparable performance on CLAHE and CMY inputs, confirming CMY as an efficient pre-processing alternative that also enhances lesion boundary sharpness. However, due to pretrained models’ bias toward RGB statistics, the best performance was observed with RGB inputs. This suggests that targeted training on CMY data could improve accuracy while retaining pre-processing efficiency, motivating future research in this direction.

\section{RESULTS and ANALYSIS}
\label{sec:res}
We evaluated five models, DenseNet201, EfficientNetB5, Vision Transformer (ViT), Swin Transformer, and the proposed DinoV2 Large, across four image pre-processing methods: RGB, CMY, CLAHE, and original images. This comparison assesses model performance on skin disease classification using accuracy and F1-score metrics.

\begin{table}[ht]
\centering
\caption{Performance of models across different image pre-processing techniques}
\label{tab:performance_comparison}
\resizebox{\linewidth}{!}{%
\begin{tabular}{|l|l|c|c|l|}
\hline
\textbf{Image Type} & \textbf{Model} & \textbf{Accuracy} & \textbf{F1-Score} & \textbf{Technique} \\ \hline

\multirow{5}{*}{RGB} 
  & DenseNet201        & 0.85 & 0.86 & Transfer Learning \\
  & EfficientNetB5     & 0.86 & 0.86 & Transfer Learning \\
  & ViT                & 0.88 & 0.89 & Transformer        \\
  & Swin Transformer   & 0.92 & 0.93 & Transformer        \\
  & \textbf{DinoV2 Large} & \textbf{0.93} & \textbf{0.93} & Transformer \\ \hline

\multirow{5}{*}{CMY} 
  & DenseNet201        & 0.82 & 0.82 & Transfer Learning \\
  & EfficientNetB5     & 0.83 & 0.84 & Transfer Learning \\
  & ViT                & 0.84 & 0.85 & Transformer        \\
  & Swin Transformer   & 0.90 & 0.92 & Transformer        \\
  & \textbf{DinoV2 Large} & \textbf{0.92} & \textbf{0.92} & Transformer \\ \hline

\multirow{5}{*}{CLAHE} 
  & DenseNet201        & 0.82 & 0.82 & Transfer Learning \\
  & EfficientNetB5     & 0.87 & 0.87 & Transfer Learning \\
  & ViT                & 0.87 & 0.88 & Transformer        \\
  & Swin Transformer   & 0.91 & 0.92 & Transformer        \\
  & \textbf{DinoV2 Large} & \textbf{0.92} & \textbf{0.93} & Transformer \\ \hline

\multirow{5}{*}{Original} 
  & DenseNet201        & 0.42 & 0.36 & Transfer Learning \\
  & EfficientNetB5     & 0.86 & 0.86 & Transfer Learning \\
  & ViT                & 0.77 & 0.73 & Transformer        \\
  & Swin Transformer   & 0.82 & 0.77 & Transformer        \\
  & DinoV2 Large       & 0.84 & 0.83 & Transformer \\ \hline

\end{tabular}
}
\end{table}

Table~\ref{tab:performance_comparison} shows the classification performance of five models evaluated on four image pre-processing pipelines: RGB, CMY, CLAHE, and unprocessed images. Among all configurations, the proposed DinoV2 Large model achieved the highest performance with an accuracy and F1-score of 0.93 on RGB images. Its performance remained stable across CMY and CLAHE variants (F1-score = 0.92–0.93), indicating strong generalization to contrast-enhanced inputs.

Traditional CNNs such as DenseNet201 exhibited lower performance overall, particularly on raw, unprocessed images (F1-score = 0.36), underscoring their sensitivity to input quality and limited capacity to generalize without explicit enhancement. In contrast, transformer-based models (e.g., ViT and Swin Transformer) demonstrated significant improvements, particularly on RGB and CLAHE images, validating their superior feature extraction and spatial modeling capabilities. Swin Transformer consistently approached DinoV2’s performance but remained marginally lower across all pre-processing methods.

From an ablation perspective, results reveal that image pre-processing has a substantial effect on performance across all architectures. While RGB inputs yielded the best outcomes overall, CMY and CLAHE provided similar gains when used with robust architectures, especially DinoV2 and Swin Transformer. Given CMY’s computational efficiency relative to CLAHE, its comparable performance (F1-score = 0.92) makes it a compelling alternative for deployment in real-time or resource-constrained environments.

\begin{figure} [ht]
\hfill
\subfloat[Original]{\label{subfig:org} \includegraphics[width=0.20\textwidth]{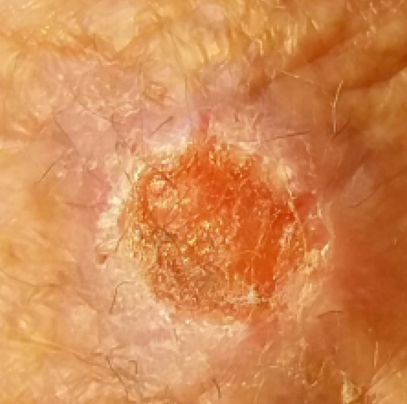}}%
\hfill
\subfloat[Heatmap]{\label{subfig:heatmap} \includegraphics[width=0.20\textwidth]{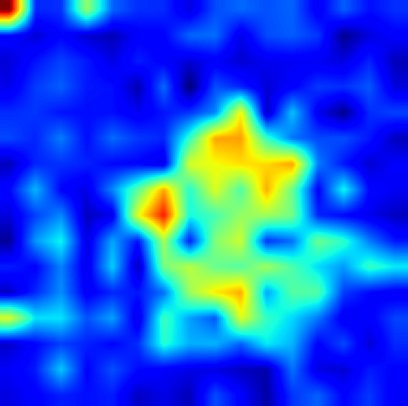}}%
\hfill
\caption{DinoV2-Large Model's Grad-CAM Visualization (a) Original, and (b) Heatmap.}
\label{fig:gradcam}
\end{figure}

Finally, to enhance model interpretability, Grad-CAM (Figure~\ref{fig:gradcam}) visualizations were employed to highlight salient regions guiding the model’s predictions. These visual explanations further corroborate DinoV2’s ability to attend to diagnostically relevant areas across pre-processing variants.

\section{CONCLUSION}
\label{sec:conc}
In this study, we developed and evaluated a deep learning framework for multi-class skin disease classification, focusing on the influence of image pre-processing techniques and model architecture on diagnostic accuracy and interpretability. Our comparative analysis of standard RGB, CMY-transformed, and CLAHE-enhanced images across both pre-trained convolutional neural networks and vision transformers revealed that CLAHE pre-processing significantly improves classification performance. Transformer-based models consistently outperformed CNNs, especially in handling visually similar and complex lesion types, by leveraging their ability to model long-range dependencies. Furthermore, Grad-CAM visualizations demonstrated that models trained on CLAHE-enhanced images exhibited improved focus on clinically relevant lesion regions, supporting the explainability and reliability of predictions. These findings underscore the importance of selecting appropriate pre-processing strategies and leveraging advanced architectures to enhance the robustness, fairness, and trustworthiness of computer-aided dermatological systems.

In the future, we would like to explore domain-specific augmentation techniques, multi-modal data fusion (e.g., combining clinical metadata with images), and the development of lightweight, deployable models for real-world applications in low-resource clinical settings.

\bibliographystyle{IEEEtran} 
\bibliography{refs}
\end{document}